  \documentclass[final]{cvpr}

\usepackage{booktabs}
\usepackage{subcaption}

\usepackage{nicefrac}
\usepackage{eucal}

\usepackage{epsfig}
\usepackage{graphicx}
\usepackage{amsmath}
\usepackage{amssymb,xspace}
\usepackage{mathtools}

\usepackage{cite}

\def\varnothing{\emptyset}

\usepackage[pagebackref=true,breaklinks=true,colorlinks,bookmarks=false]{hyperref}

\usepackage[margin=4pt,font=small,labelfont=bf,labelsep=endash,tableposition=top]{caption}

\def\Ours{{VisTR}\xspace}
\newcommand{\myparagraph}[1]{{\vspace{0.01cm} \noindent \bf #1}}

\begin{document}
\pagenumbering{gobble}
\title{End-to-End Video Instance Segmentation with Transformers}

\author{
Yuqing Wang$^1$, 
Zhaoliang Xu$^1$, 
Xinlong Wang$^2$,
Chunhua Shen$^2$, 
Baoshan Cheng$^1$,
Hao Shen$^1$\thanks{Corresponding author.}, 
Huaxia Xia$^1$
\\[0.2cm] 
$ ^1$ Meituan 
 ~
 ~
 ~
 ~
 ~
$ ^2$ The University of Adelaide, Australia
\\{\tt\small yuqingwang1029@gmail.com, shenhao04@meituan.com}
}

\makeatletter
\let\@oldmaketitle\@maketitle
\renewcommand{\@maketitle}{\@oldmaketitle
 \centering
    \includegraphics[width=.7681\textwidth]{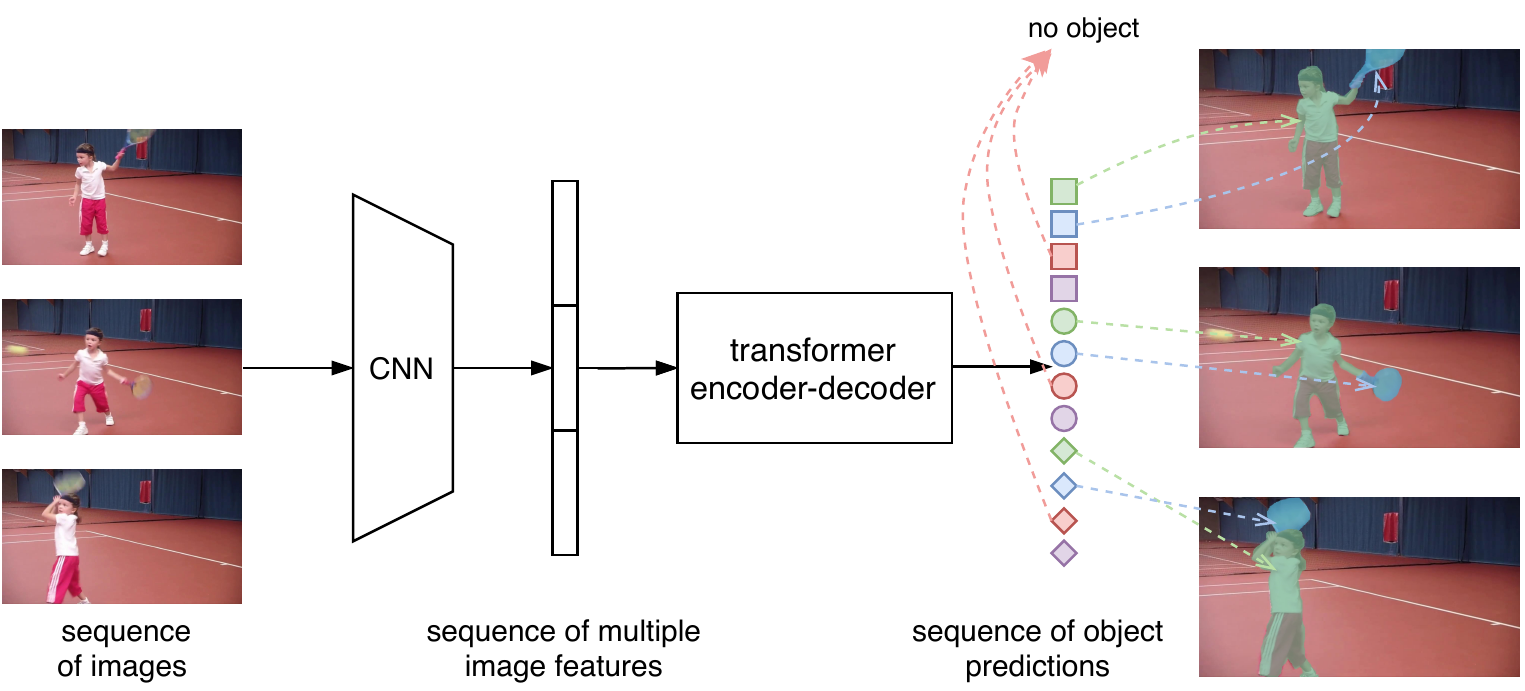}
    \vspace{-0.21cm}
    \captionof{figure}{
    \textbf{Overall pipeline of \Ours.} 
The model takes a sequence of images as input and outputs a sequence of 
instance predictions. 
Here same shapes represent predictions in one image, and same colors represent predictions of the same object instance. Note that the overall predictions follow the input frame order, and the order of object predictions for different images keeps the same (Best viewed on screen). 
    }
    \label{fig:1}
  \bigskip}
\makeatother
\maketitle

\begin{abstract}

Video instance segmentation (VIS) is the task that requires simultaneously classifying, segmenting and tracking object instances of interest in video. 
Recent methods typically develop sophisticated pipelines to tackle this task. 
Here, we propose a new video instance segmentation framework built upon Transformers, termed {\rm \Ours}, which views the VIS task as a direct end-to-end parallel sequence decoding/prediction problem. Given a video clip consisting of multiple image frames as input, 
\Ours outputs the sequence of masks for each instance in the video in order directly.
At the core is
a new, effective instance sequence matching and segmentation strategy, which supervises and segments instances at the sequence level as a whole.  
\Ours frames the instance segmentation and tracking in the same perspective of similarity learning, thus considerably simplifying the overall pipeline and is significantly different from existing approaches.

Without 
bells and whistles, \Ours achieves the highest speed among all existing VIS models, and achieves the best result 
among methods using single model on the YouTube-VIS dataset. 
For the first time, we demonstrate a much simpler and faster video instance segmentation framework built upon Transformers, 
achieving competitive accuracy. 
We hope that \Ours  can motivate future research for more video understanding tasks. 

Code is available at:
\def\UrlFont{\tt \color{blue}}
\url{https://git.io/VisTR}

\end{abstract}

\section{Introduction}

Instance segmentation is one of the fundamental tasks in computer vision.  
While significant progress has been witnessed in instance segmentation of images
\cite{he2017mask,tian2020conditional,wang2020solo,SoloV22020,chen2020blendmask,wang2020centermask}, 
much less effort was spent on segmenting instances in videos. 
Here we propose a new video instance segmentation framework built upon Transformers. 
Video instance segmentation (VIS), recently proposed in \cite{vis2019}, 
requires one to simultaneously classify, 
segment and track object instances of interest in a video sequence. 
It is more challenging in that one needs 
to perform instance segmentation for each individual frame and at the same time to establish data association of instances across consecutive frames, \textit{a.k.a.},  tracking.

State-of-the-art methods typically develop sophisticated pipelines to tackle this task. 
Top-down approaches~\cite{vis2019,bertasius2020classifying} follow the tracking-by-detection paradigm, relying heavily on image-level instance segmentation models \cite{he2017mask,chen2019hybrid} and complex human-designed rules to associate the instances. 
Bottom-up approaches \cite{Athar_Mahadevan20ECCV} separate object instances by clustering learned pixel embeddings. 
Due to heavy reliance on the dense prediction quality, these methods often need multiple steps to generate the masks iteratively, which makes them slow. 
Thus, a simple, end-to-end trainable VIS framework is highly desirable.

    Here, we take a deeper look at the video instance segmentation task. 
    Video frames contain richer information than single images such as motion patterns and temporal consistency of instances, offering useful cues for instance segmentation, and classification.
    At the same time, the better learned instance features can help tracking of instances. In essence, the instance segmentation and instance tracking are both concerned with similarity learning: 
    instance segmentation is to learn the pixel-level similarity and instance tracking is to learn the similarity between instances. 
    Thus, it is natural to solve these two sub-tasks in a single framework and benefit each other.
Here we aim to develop such an end-to-end VIS framework. The framework needs to be simple and achieves strong performance without whistles and bells.
To this end, 
we propose to %
employ
the Transformers \cite{vaswani2017attention}. 
Importantly, for the first time we demonstrate that, as the Transformers provide 
 building blocks, \textit{it enables one to design a simple and clean framework for VIS, and possibly for a much wider range of video processing tasks in computer vision}.  Thus potentially, it is possible to unify most vision tasks of different input modalities---such as image, video and point clouds processing---into the Transformer framework.
Transformers are widely used for sequence to sequence learning in NLP \cite{vaswani2017attention}, and start to show promises in vision \cite{Detr,dosovitskiy2020}.
Transformers are capable of modeling long-range dependencies, 
and thus can be naturally applied to video for learning temporal information across multiple frames. 
In particular, the core mechanism of Transformers, self-attention, is designed to learn and update the features based on all pairwise similarities between them. 
The above characteristics of Transformers make them great candidates for the VIS task.

In this paper, we propose the Video Instance Segmentation TRansformer (\Ours), which views the VIS task as a parallel sequence decoding/prediction problem. 
Given a video clip that consists of multiple image frames as input, the \Ours outputs the sequence of masks for each instance in the video in order directly. The output sequence for each instance is referred to as \textit{instance sequence} in this paper. 
The overall \Ours pipeline is illustrated in \figref{fig:1}. In the first stage, given a sequence of video frames, 
a standard CNN module extracts features of individual image frames, then the multiple image features are concatenated in the frame order to form the clip-level feature sequence. In the second stage, the Transformer takes the clip-level feature sequence as input, and outputs a sequence of object predictions in order. In 
\figref{fig:1}
same shapes represent predictions for the same image, and the same colors
represent the same instance of different images. The sequence of predictions follow the order of input images, and the predictions of each image follows the same instance order. 
Thus, instance tracking is achieved \textit{seamlessly and naturally} in the same framework of instance segmentation.

To achieve this goal, there are two main challenges: 1) how to maintain the order of outputs and 2)  how to obtain the mask sequence for each instance out of the Transformer network. Correspondingly, we introduce the \textit{instance sequence matching} strategy and the \textit{instance sequence segmentation} module. The instance sequence matching performs bipartite graph
matching between the output instance sequence and the ground-truth instance sequence, and supervises the sequence as a whole.
Thus, the order can be maintained directly. The instance sequence segmentation accumulates the mask features for each instance across multiple frames through self-attention and segments the mask sequence for each instance through 3D convolutions. 

%

Our main contributions are summarized as follows.
\begin{itemize}
\itemsep -0.17cm
    \item 
    We propose a new video instance segmentation framework built upon Transformers, 
    termed
    \textbf{\Ours}, 
    which 
    views the VIS task as a \textit{direct end-to-end} 
    parallel 
    sequence decoding/prediction problem. The framework is significantly different from existing approaches, considerably simplifying the overall pipeline.
    \item 
    \Ours solves the %
    VIS
    from a new perspective of similarity learning. Instance segmentation is to learn the pixel-level similarity and instance tracking is to learn the similarity between instances. %
    Thus, 
    instance tracking is achieved \textit{seamlessly and naturally} in the same framework of instance segmentation.
    \item
    The key to the success of \Ours is 
    a new strategy for \textit{instance sequence matching and segmentation},
    which is tailored for our framework. 
    This carefully-designed strategy enables us to 
    supervise and segment instances at the sequence level as a whole.  
    
    \item
    \Ours achieves strong 
    results on the YouTube-VIS dataset, achieving  40.1\% in mask mAP at the speed of 57.7 FPS  
    , which is the best and fastest among methods that use a single model.

\end{itemize}

\section{Related work}

\myparagraph{Video object segmentation.} VOS \cite{DAVIS16} is closely related to VIS.
Analogue to  object tracking, which is detecting boxes of foreground objects 
in a class-agnostic fashion, VOS is segmenting masks of foreground 
class-agnostic objects. 
Same as in tracking, usually one is allowed to use only the first few frames' annotations for training. In contrast, VIS requires to segment and track all instance masks of a fixed category set of objects in a video sequence.

\myparagraph{Video instance segmentation.}
The VIS task \cite{vis2019} requires classifying, segmenting instances in each frame and linking the same instance across frames. State-of-the-art methods typically develop sophisticated pipelines to tackle it. MaskTrack R-CNN\cite{vis2019} extends the Mask R-CNN \cite{he2017mask} with a tracking branch and external memory that saves the features of instances across multiple frames. Maskprop \cite{bertasius2020classifying} builds on the Hybrid Task Cascade Network \cite{chen2019hybrid}, and re-uses the predicted masks to crop the extracted features, then propagates them temporally to improve the segmentation and tracking. STEm-Seg \cite{Athar_Mahadevan20ECCV} proposes to model video clips as 3D space-time volumes and then separates object instances by clustering learned embeddings. 
Note that the above approaches either rely on complex heuristic rules to associate the instances or require multiple steps to generate and optimize the masks iteratively.  
In contrast, here 
we aim to build a \textit{simple and end-to-end trainable} VIS framework.

\myparagraph{Transformers.}
Transformers were first proposed in \cite{vaswani2017attention} for the sequence-to-sequence machine translation task, and since then have  become the \textit{de facto}
method in most NLP tasks. 
The core mechanism of Transformers, self-attention, makes it particularly suitable for modeling long-range dependencies. 
Very recently, 
Transformers start to show promises in solving computer vision tasks. DETR \cite{Detr} builds an object detection systems based on Transformers, which largely simplifies the traditional detection pipeline, and achieves \textit{on par}
performances compared with highly-optimized CNN based detectors\cite{ren2015faster}. Our work here is inspired by DETR.  
ViT \cite{dosovitskiy2020} introduces the Transformer to image recognition and models an image as a sequence of patches, which attains excellent results compared to state-of-the-art convolutional networks. The above works show the effectiveness of Transformers in image understanding tasks. 
To our knowledge, thus far there are 
no prior applications of Transformers to video instance segmentation. 
It is intuitive to see that the Transformers' advantage of modeling long-range dependencies makes it an ideal candidate for learning temporal information across multiple frames for video understanding tasks. Here, 
we propose the \Ours method and provide an affirmative answer to that.
As the original Transformers are auto-regressive models, which generate output tokens one by one, 
for efficiency, \Ours employs a non-auto-regressive variant of the Transformer to achieve parallel sequence generation.

\section{Our  Method: \Ours}

\begin{figure*}[h]
\centering
\includegraphics[width=.95\linewidth]{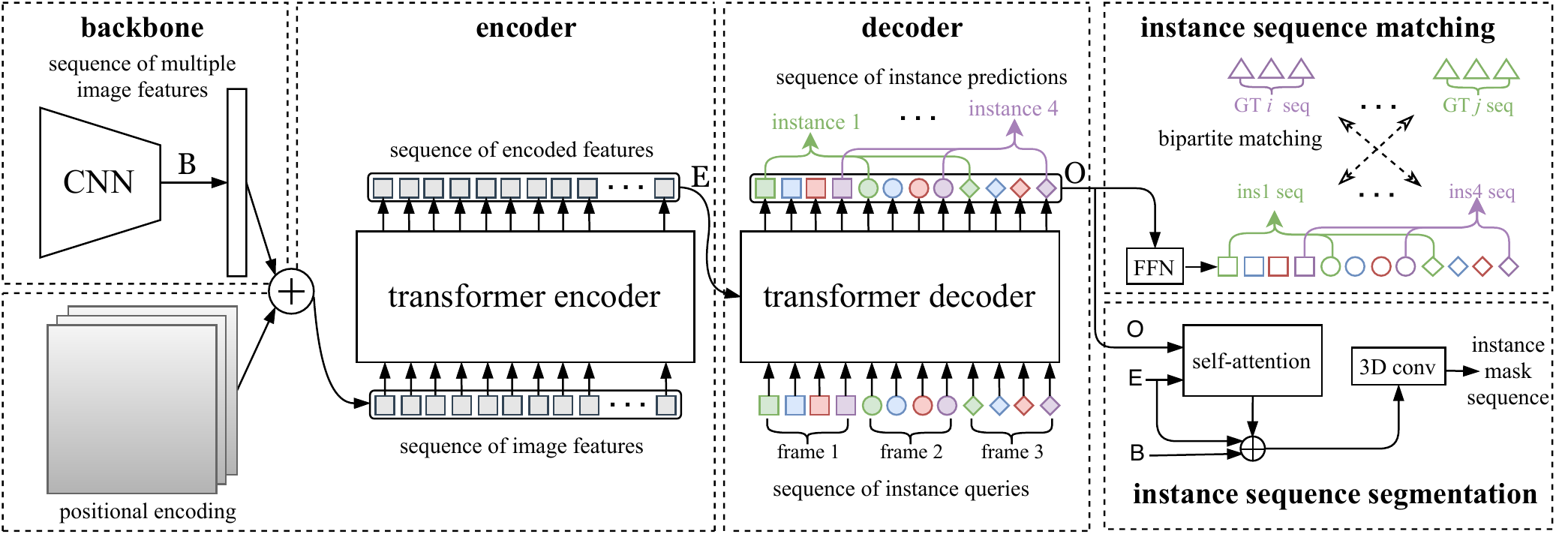}
\caption{\textbf{The overall architecture of \Ours.} It contains four main components: 1) 
a CNN backbone that extracts feature representation of multiple images;
2) an encoder-decoder Transformer that models the relations of pixel-level features and decodes the instance-level features; 3) an instance sequence matching module 
that supervises the model; and 4) an instance sequence segmentation module that outputs the final mask sequences (Best viewed on screen).
}
\label{fig:2}
\end{figure*}

We tackle the video instance segmentation task by modeling it as a direct sequence prediction problem.
Given a video clip that consists of multiple image frames as input, the \Ours outputs the sequence of masks for each instance in the video in order. 
To achieve this goal, we introduce the instance sequence matching and segmentation strategy to supervise and segment the instances at the sequence level as a whole. 
In this section, we first introduce the overall architecture of the proposed \Ours in \secref{sec:3.1}, then the details of the instance sequence matching and segmentation module will be described in \secref{sec:3.2} and \secref{sec:3.3} respectively.  

\subsection{\Ours Architecture}\label{sec:3.1}
The overall \Ours architecture is depicted in \figref{fig:2}. It contains four main components: a CNN backbone to extract compact feature 
representations
of multiple frames,
an encoder-decoder Transformer to model the similarity of pixel-level and instance-level features, an instance sequence matching module for supervising the model, and an instance sequence segmentation module.

\myparagraph
{Backbone.} 
The backbone %
extracts the original pixel-level feature sequence of the input video clip.
Assume that the initial video clip with $T$ frames of resolution $H_{0}\times W_{0}$  is denoted by $x_{clip}\in\mathbb{R}^{T\times 3 \times H_{0}\times W_{0}}$.
First, a standard CNN backbone generates a lower-resolution activation map for each frame, then the features for each frame are concatenated to form the clip level feature map $f_{0}\in\mathbb{R}^{T\times C \times H\times W}$.

\myparagraph
{Transformer encoder.}
The Transformer encoder is 
employed 
to model the similarities among all the pixel level features in the clip.
First, a 1$\times$1 convolution is applied to the above feature map, reducing the dimension from $C$ to %
$d$ ($ d < C$), %
resulting in 
a new feature map $f_{1}\in\mathbb{R}^{T\times d \times H\times W}$.
To form a clip level feature sequence that can be fed into the Transformer encoder, we flatten the spatial and temporal dimensions of $f_{1}$ into one dimension, resulting in a 2D feature map of size $d\times ( T \cdot H \cdot W )$.
Note that the temporal order is always in accordance with that of the initial input. Each encoder layer has a standard architecture that consists of a multi-head self-attention module and a fully connected feed forward network (FFN). 

\myparagraph
{Temporal and spatial positional encoding.} The Transformer architecture is permutation-invariant, while the segmentation task requires precise position information. To compensate for this, we supplement the features with fixed positional encodings information that contains the three dimensional (temporal, horizontal and vertical) positional information in the clip. 
Here we adapt the positional encoding in the original Transformer \cite{vaswani2017attention} for our 3D case.
Specifically, for the coordinates of each dimension we independently use $\nicefrac{d}{3}$ sine and cosine functions with different frequencies:
\def\PE{{\rm PE}}
\def\pos{{\rm pos}}
\begin{equation}
%
\PE(\pos,i)=
\begin{cases}
 \sin \bigl( \pos \cdot \omega_{k} \bigr),  ~~~~\text{for } i=2k, 
\\
\cos \bigl( \pos \cdot \omega_{k} \bigr),   ~~~~\text{for } i=2k+1;
\end{cases}
\label{eq:pe}
\end{equation}
where $ \omega_{k}= 1/ 10000^{2k/ \frac{d}{3}}$;
`$\pos$' is the position in the corresponding dimension. Note that the $d$ should be divisible by 3, as the positional encodings of the three dimensions should be concatenated to form the final $d$ channel positional encoding. These encodings are added to the input of each attention layer. 
\myparagraph
{Transformer decoder.} 
The Transformer decoder aims to decode the top pixel features that can represent the instances of each frame, which is called instance level features. Motivated by DETR \cite{Detr}, we also introduce a
fixed number of input embeddings to query the instance features from pixel features, termed as \textit{instance queries}. Suppose that the model decodes $n$ instances each frame, then for $T$ frames the instance query number is $N=n \cdot  T$. The instance queries are learned by the model and have the same dimension with the pixel features. 
Taking the output of encoder $E$ and $N$ instance queries $Q$ as input, the Transformer decoder outputs $N$ instance features, denoted by $O$ in \figref{fig:2}. The overall predictions follow the input frame order, and the order of instance predictions for different images is the same. 
Thus, the tracking of instances in different frames could be realized by linking the items of the corresponding indices directly. 
\subsection{Instance Sequence Matching}\label{sec:3.2}

\Ours infers a fixed-size sequence of $N$ predictions, in a single pass through the decoder. One of the main challenges for this framework is to maintain the relative position of predictions for the same instance in different images, \textit{a.k.a.}, \textit{instance sequence}. In order to find the corresponding ground truth and supervise the instance sequence as a whole, we introduce the instance sequence matching strategy.

As the \Ours decode $n$ instances each frame, the number of instance sequence is also $n$. Let us denote by $\hat y=\left \{{ \hat y_{i}} \right \}_{i=1}^{n}$ the predicted instance sequences, and $y$ the ground truth set of instance sequences. Assuming $n$ is larger than the number of instances in the video clip, we consider $y$ also as a set of size $n$ padded with $\varnothing$. In order to find a bipartite 
graph
matching between the two sets, we search for a permutation of $n$ elements $\sigma \in S_{n}$ with the lowest cost:
\begin{equation}
\hat{\sigma}=\underset{\sigma \in S_{n}}{\arg \min } \sum_{i}^{n} \mathcal{L}_{\text{match}}\left(y_{i}, \hat{y}_{\sigma(i)}\right)
\label{eq:1}
\end{equation}
where $\mathcal{L}_{\operatorname{match}}\left(y_{i}, \hat{y}_{\sigma(i)}\right)$ is a pair-wise \textit{matching cost} between ground truth $y_{i}$ and an instance sequence prediction with index $\sigma(i)$. The optimal assignment could be computed efficiently by the Hungarian algorithm\cite{kuhn1955hungarian}, following prior work (\textit{e.g.}, \cite{stewart2016end}).

As computing the mask sequence similarity directly is computationally intensive, we find a surrogate, the box sequence to perform the matching. To obtain the box predictions, we apply a 3-layer feed forward network (FFN) with ReLU activation function and a linear projection layer to the object predictions $O$ of Transformer decoder. Following the same practice of DETR \cite{Detr}, the FFN 
predicts the normalized center coordinates, height and width of the box w.r.t.\  input image, and the linear layer predicts the class label using a softmax function. We also add a ``background" class to represent that no object is detected. 

Given the $N=n \cdot  T$ bounding box predictions for the object predictions sequence, we could associate $n$ box sequences for each instance by their indices, referred to as ins1 box seq...ins4 box seq in \figref{fig:2}. The matching loss takes both the class predictions and the similarity of predicted and ground truth boxes into account. Each element $i$ of the ground truth set can be seen as 
\begin{equation}
y_{i} = \{
      (c_{i},c_{i}  ...,c_{i}),  (b_{i,0},b_{i,1}...,b_{i,T}) 
         \}
\end{equation}
where $c_{i}$ is the target class label (which may be $ \varnothing$) for this instance, and $b_{i,t}\in[0,1]^{4}$ is a vector that defines ground truth box center coordinates and its relative height and width in the frame $t$. $T$ represent the number of input frames. Thus, for the predictions of instance with index $\sigma(i)$ we denoted the probability of class $c_{i}$ as  
\begin{equation}
{\hat p}_{(\sigma(i))}(c_{i})  =  \{{\hat p}_{(\sigma(i),0)} (c_{i})...,\hat p_{(\sigma(i),T)}(c_{i}) \}
\end{equation}
and the predicted box sequence as 
\begin{equation}
\hat b_{\sigma(i)}= \left \{ \hat b_{(\sigma(i),0)},\hat b_{(\sigma(i),1)}..., \hat b_{(\sigma(i),T)} \right \}
\end{equation}
With the above notation, we define
\begin{equation}
\mathcal{L}_{\text{match}}\left(y_{i}, \hat{y}_{\sigma(i)}\right)=- \hat{p}_{\sigma(i)}\left(c_{i}\right)+ \mathcal{L}_{\mathrm{box}}
\bigl(b_{i}, \hat{b}_{\sigma(i)}
\bigr),
\label{eq:2}
\end{equation}
where $c_{i} \neq \varnothing$.
Based on the above criterion, we could find the one-to-one matching of the sequences by the Hungarian algorithm. Given the optimal assignment, we could compute the loss function, the \textit{Hungarian loss} for all pairs matched in the previous step. The loss is a linear combination of a negative log-likelihood for class prediction, a box loss and mask loss for the instance sequences:
\begin{align}
\mathcal{L}_{\text{Hung}}(y, \hat{y})  
&= 
\sum_{i=1}^{N} \Bigl[ 
    ( -\log \hat{p}_{\hat{\sigma}(i)}(c_{i})) 
+ \mathcal{L}_{\text {box }}(b_{i}, \hat{b}_{\hat{\sigma}}(i)) 
\notag
\\ 
&+
\mathcal{L}_{\text {mask }}(m_{i}, \hat{m}_{\hat{\sigma}}(i)) 
    \Bigr].
\label{eq:3}
\end{align}
Here $c_{i} \neq \varnothing$,
and $\hat \sigma$ is the optimal assignment computed in Eq.~(\ref{eq:1}). The \textit{Hungarian loss} is used to train the whole framework. 

The second part of the matching cost and the Hungarian loss is $\mathcal{L}_{\rm box}$ that scores the bounding boxes. We use a linear combination of the sequence level $\mathcal{L}_{1}$ loss and the generalized IOU loss \cite{rezatofighi2019generalized}: 
\begin{align}
\mathcal{L}_{\mathrm{box}}\bigl(b_{i}, \hat{b}_{\sigma(i)} 
\bigr)
& =
\frac{1}{T} \sum_{t=1}^{T}   
\Bigl[ 
    \lambda_{\mathrm{iou}}  \cdot   \mathcal{L}_{\mathrm{iou}}\bigl( b_{i,t}, \hat{b}_{\sigma(i),t} \bigr) 
\notag 
\\ 
& +
\lambda_{\mathrm{L} 1}\left\|b_{i,t}-\hat{b}_{\sigma(i), t}\right\|_{1} 
\Bigr]. 
\label{eq:4}
\end{align}
Here $\lambda_{\mathrm{iou}}, \lambda_{\mathrm{L} 1} \in \mathbb{R}$ are hyper-parameters. These two losses are normalized by the number of instances inside the batch.
In the sequel, we present the details.

\subsection{Instance Sequence Segmentation}\label{sec:3.3}

The instance sequence segmentation module aims to predict the mask sequence for each instance. To realize that, the model needs to accumulate the mask features of multiple frames for each instance firstly, then the mask sequence segmentation is performed on the accumulated features. 

The mask features are obtained by computing the similarity map between the object predictions $O$ and the Transformer encoded features $E$. To simplify the calculation, we only compute with the features of its corresponding frame for each object prediction. For each frame, the object predictions $O$ and the corresponding encoded feature maps $E$ are fed into the self-attention module to obtain the initial attention maps. Then the attention maps will be fused with the initial backbone features $B$ and the transformed encoded features $E$ of the corresponding frames, following a similar practice with the DETR\cite{Detr}. The last layer of the fusion is a deformable convolution layer\cite{dai2017deformable}. In this way, the mask features for each instance of different frames are obtained.

Following the same spirit of taking the instance sequence as a whole, the mask features of the same instance in different frames should be propagated and reinforce each other. We propose to utilize the 3D convolution to realize that. 
Assume that the mask feature for instance $i$ of frame $t$ is $g_{i,t}\in \mathbb{R}^{1 \times a \times \nicefrac{H_{0}}{4} \times \nicefrac{W_{0}}{4}}$, where $a$ is the channel number, then we concatenate the features of $T$ frames to form the $G_{i}\in \mathbb{R}^{1 \times a\times T \times \nicefrac{H_{0}}{4} \times \nicefrac{W_{0}}{4}}$.
The \textit{instance sequence segmentation} module takes the instance sequence mask feature $G_{i}$ as input, and output the mask sequence $m_{i}\in \mathbb{R}^{1 \times 1\times T \times \nicefrac{H_{0}}{4} \times \nicefrac{W_{0}}{4}}$ for the instance directly. This module contains three 3D convolutional layers and Group Normalization\cite{wu2018group} layers with ReLU activation function. No normalization or activation is performed after the last convolution layer, and the output channel number of the last layer is 1. In this way, the masks of the instance for $T$ frames are obtained. The mask loss for supervising the predictions in Eq.~\eqref{eq:3} is defined as a combination of the Dice \cite{milletari2016v} and Focal loss \cite{Lin_2017_ICCV}:
\def\Dice{{ {\cal L}_{ \text{Dice}}} }
\def\Focal{{ {\cal L}_{ \text{Focal}}} }
\begin{align}
\mathcal{L}_{\text{mask}}\left(m_{i}, \hat{m}_{\sigma(i)}\right)
& =
\lambda_{\text{mask}}\frac{1}{T}\sum_{t=1}^{T} 
\Bigl[ 
 \Dice(m_{i,t}, \hat{m}_{\sigma(i),t}
) 
\notag
\\
&+
\Focal(m_{i,t}, \hat{m}_{\sigma(i),t}) 
\Bigr].  
\label{eq:5}
\end{align}

\section{Experiments}

In this section, we conduct experiments on the YouTube-VIS \cite{vis2019} dataset, which contains 2238 training, 302 validation and 343 test video clips. Each video of the dataset is annotated with per pixel segmentation mask, category and instance labels. The object category number is 40. As the test set evaluation is closed, we evaluate our method in the validation set. The evaluation metrics are average precision (AP) and average recall (AR), with the video Intersection over Union (IoU) of the mask sequences as the threshold.

\subsection{Implementation Details}
%
\myparagraph{Model settings.} 
As the largest number of the annotated video length for YouTube-VIS \cite{vis2019} is 36, we take this value as the default input video clip length $T$.
Thus, no post-processing is needed to associate different clips from one video, which makes our model totally end-to-end trainable. The model predicts 10 objects for each frame, thus the total object query number is 360. For the Transformer we use 6 encoder, 6 decoder layers of width 384 with 8 attention heads. Unless otherwise specified, ResNet-50 \cite{he2016deep} is used as our backbone networks and the same hyper-parameters of 
DETR \cite{Detr} are used.

\myparagraph{Training.} The model is implemented with PyTorch-1.6 \cite{paszke2019pytorch}, trained with AdamW \cite{loshchilov2018decoupled} of initial Transformer’s learning rate being $10^{-4}$ , the backbone’s learning rate being $10^{-5}$. The models are trained for 18 epochs, with the learning rate decays by 10x at 12 epochs. We initialize our backbone networks with the weights of DETR pretrained on COCO \cite{lin2014microsoft}. The models are trained on 8  V100 GPUs of 32G RAM, with 1 video clip per GPU. The frame sizes are downsampled to 300$\times$540 to fit the GPU memory.

\myparagraph{Inference.} During inference, we follow the same scale setting as training. No post-processing is needed for associating instances. Instances with scores larger than 0.001 are kept. The mean score for all the frames is used as the instance score. For instances that have been classified to different categories in different frames, we use the most frequently predicted category as the final instance category.

\subsection{Ablation Study}

In this section we conduct extensive ablation experiments to study the core factors of \Ours.
Comparison results are 
reported
in Table \ref{tab:1}.

\begin{table*}[!t]
\small
\begin{subtable}{0.5\linewidth}
\centering
\captionsetup{width=0.9\linewidth}
\begin{tabular}{c|c|cccc}
Length & AP & $\rm AP_{50}$ & $\rm AP_{75}$ & $\rm AR_{1}$ & $\rm AR_{10}$ \\
\hline
18&29.7&50.4&31.1&29.5&34.4\\
24&30.5&47.8&33.0&29.5&34.4\\
30&31.7&53.2&32.8&31.3&36.0\\
36&33.3&53.4&35.1&33.1&38.5
\end{tabular}
\caption{\textbf{Video sequence length.} 
The performance improves as the sequence length increases.
}
\label{tab:length}
\end{subtable}%
\begin{subtable}{0.5\linewidth}
\centering
\captionsetup{width=0.9\linewidth}
\setlength{\tabcolsep}{1mm}{
\begin{tabular}{ r |c|c|cccc}
 & 
 \# 
 &AP & $\rm AP_{50}$ & $\rm AP_{75}$ & $\rm AR_{1}$ & $\rm AR_{10}$ \\
\hline
video level& 1&8.4&13.2&9.5&20.0&20.8\\
frame level& 36&13.7&23.3&14.5&30.4&35.1\\
ins. level& 10&32.0&52.8&34.0&31.6&37.2\\
pred. level& 360&33.3&53.4&35.1&33.1&38.5
\end{tabular}
}
\caption{\textbf{Instance query embedding}. 
Instance-level query is only 1.3\%  lower in AP than the prediction-level query with 36$\times$ fewer embeddings. }
\label{tab:query}
\end{subtable}%

\begin{subtable}{0.5\linewidth}
\centering
\captionsetup{width=0.9\linewidth}
\begin{tabular}{c|c|cccc}
time order & AP & $\rm AP_{50}$ & $\rm AP_{75}$ & $\rm AR_{1}$ & $\rm AR_{10}$ \\
\hline
random&32.3&52.1&34.3&33.8&37.3\\
in order&33.3&53.4&35.1&33.1&38.5
\end{tabular}
\caption{\textbf{Video sequence order.} Sequence in time order is 1.0\%  better  in AP than sequence in random order.}
\label{tab:order}
\end{subtable}%
\vspace{-0.3cm}
\begin{subtable}{0.5\linewidth}
\centering
\captionsetup{width=0.9\linewidth}
\begin{tabular}{c|c|cccc}
 & AP & $\rm AP_{50}$ & $\rm AP_{75}$ & $\rm AR_{1}$ & $\rm AR_{10}$ \\
\hline
w/o&28.4&50.1&29.5&29.6&33.3\\
w &33.3&53.4&35.1&33.1&38.5
\end{tabular}
\caption{\textbf{Position encoding}. Position encoding brings about 5\% AP gains to \Ours.}
\label{tab:pos}
\end{subtable}%

\begin{subtable}{0.5\linewidth}
\centering
\captionsetup{width=0.9\linewidth}
\begin{tabular}{c|c|cccc}
 & AP & $\rm AP_{50}$ & $\rm AP_{75}$ & $\rm AR_{1}$ & $\rm AR_{10}$ \\
\hline
CNN &32.0&54.5&31.5&31.6&37.7\\
Transformer&33.3&53.4&35.1&33.1&38.5
\end{tabular}
\caption{\textbf{CNN-encoded feature vs.\  Transformer-encoded feature} for mask prediction. The transformer improves the feature quality.}
\label{tab:encoded}
\end{subtable}%
\begin{subtable}{0.5\linewidth}
\centering
\captionsetup{width=0.9\linewidth}
\begin{tabular}{c|c|cccc}
& AP & $\rm AP_{50}$ & $\rm AP_{75}$ & $\rm AR_{1}$ & $\rm AR_{10}$ \\
\hline
w/o &33.3&53.4&35.1&33.1&38.5\\
w&34.4&55.7&36.5&33.5&38.9
\end{tabular}
\caption{\textbf{Instance sequence segmentation module.} The module with 3D convolutions brings 1.1\% AP gains.}
\label{tab:refine}
\end{subtable}
\caption{Ablation experiments for \Ours. All models are trained on YouTubeVIS \texttt{train} for 10 epochs and tested on YouTubeVIS \texttt{val}, using the ResNet-50 backbone.}\label{tab:1}
\vspace{-0.7cm}
\end{table*}

The main difference between video and image is that video contains  temporal information.
How to effectively learn and exploit temporal information is the key to video understanding.
Firstly, we study the importance of temporal information to \Ours in two dimensions: the amount and the order. 

\myparagraph{Video sequence length.}
To evaluate the importance of the amount of temporal information to \Ours, we experiment with models trained with different input video sequence lengths. As reported in Table~\ref{tab:length}, with the length varying 
from 18 to 36, the AP increases monotonically from 29.7\%  to 33.3\%. This result shows that more temporal information indeed helps the model learn better. As the largest video length of the dataset is 36, we argue that, if with a larger dataset, \Ours can achieve even better results.
Note that for this experiment, if the clip length is less than the video length, instance matching in overlapping frames is used for associating them from different clips.

\myparagraph{Video sequence order.}
As the movement of objects in %
real scenes 
are continuous, we %
believe 
that the order of temporal information is also important. To evaluate, we perform a comparison of the model trained with input video sequence in random order vs.\  time order. Results in Table~\ref{tab:order} show that the model learned %
according to the
time order information achieves 1 point higher, which verifies the importance of the temporal order.

\myparagraph{Positional encoding.}
Position information is important for the dense prediction problem of VIS. As the original feature sequence contains no positional information, we supplement with the spatial and temporal positional encodings,
which indicate the relative positions in the video sequence. Experiments of models with and without positional encoding are %
presented 
in Table~\ref{tab:pos}. The model without positional encoding %
manages to 
achieve 28.4\% AP.
Our explanation is that the ordered format of the sequence supervision and the correspondence between the input and output order of the Transformer provide some relative positional information \textit{implicitly}. In the second experiment, the performance improves by about 5 points, which verifies the necessity of explicit positional encoding.

\myparagraph{Instance queries.}
The instance queries are learned embeddings for decoding the representative instance predictions.
In this experiment, we study the effect of instance queries and attempt to exploit the inner connections among them by varying the embedding number. Suppose the model decode $n$ instances each frame, and the frame number is $T$. The input instance query number should be $n \times T$ to decode the same number for predictions.
In the default setting, one embedding is responsible for one prediction, the model directly learns $n \times T$ unique embeddings, termed as `prediction level' in Table~\ref{tab:query}. 
In the `video level setting', one embedding is learned for all the instance predictions, \textit{i.e.}, the same embedding is repeated $n\times T$ times as the input of decoder. In the `frame-level' setting, the model only learns $T$ unique embeddings and repeats them by $n$ times.
In the `instance level' setting, the model only learns $n$ unique embeddings and repeats them by $T$ times. 
The n and T corresponds to the value of 10 and 36 in the table respectively. The result is 8.4\% AP and 13.7\% AP for `video level' and `frame level' settings respectively.
Surprisingly, the `instance level' queries can achieve 32.0\% AP, which is only 1.3 points lower than the default setting. The result shows that the queries for one instance can be shared for the \Ours model, which makes the tracking natural. But the queries for one frame can not be shared.

\myparagraph{Transformers for feature encoding.}
As illustrated in the 
`instance sequence segmentation' module of \figref{fig:2}.
The module takes three types of features as input: 
the features `B' from the backbone, the feature `E' from the encoder and the attention map computed by the feature `E' and `O'. 
To show the superiority of Transformers in feature encoding, we compare the results of using the original input `O' vs.\ output `E' of the encoder for the second feature, \textit{a.k.a.}, CNN-encoded features vs.\  Transformer-encoded features. As reported in Table~\ref{tab:encoded}, the CNN-encoded features achieves 32.0\% AP, and the Transformer-encoded features achieve 1.3 points higher. This demonstrates that features are learned better after the Transformer updates them based on all pairwise similarities between them through self-attention. The result also shows the superiority of modeling the spatial and temporal features as a whole.

\myparagraph{Instance sequence segmentation.}
The segmentation process contains both the instance mask feature accumulation and instance sequence segmentation modules. The instance sequence segmentation module takes the instance sequence as a whole. We expect that it can strengthen the mask prediction by learning the temporal information through 3D convolutions. Thus, when objects are in challenging situations such as occlusions or motion blurs, the module can learn to propagate information from other frames to help the segmentation. Besides, the features of the same instance from multiple frames could help the network recognize the instance better. In this experiment, we perform a study of models with or without the 3D instance sequence segmentation module. For the former case, we apply a 2D convolutional layer with the output channel being 1 to the mask features for each instance of each frame to obtain the masks. The comparison is shown in Table~\ref{tab:refine}.
The instance sequence segmentation module improves the result by 1.1 points, which verifies the effectiveness of the proposed module.

With
these ablation studies, we conclude that in 
\Ours design: the temporal information, positional encodings, instance queries, global self-attention in the encoder and the instance sequence segmentation module, 
all play important roles \textit{w.r.t.}\ the final performance.

\subsection{Main Results}

\begin{table*}
\small 
\centering 
\begin{tabular}{ r|l|c|c|cccc}
 Method&backbone& FPS& AP & $\rm AP_{50}$ & $\rm AP_{75}$ & $\rm AR_{1}$ & $\rm AR_{10}$ \\
\hline
DeepSORT\cite{wojke2017simple}&ResNet-50&-&26.1&42.9&26.1&27.8&31.3\\
FEELVOS\cite{voigtlaender2019feelvos}&ResNet-50&-&26.9&42.0&29.7&29.9&33.4\\
OSMN\cite{yang2018efficient}&ResNet-50&-&27.5&45.1&29.1&28.6&33.1\\
MaskTrack R-CNN\cite{vis2019}&ResNet-50&20.0&30.3&51.1&32.6&31.0&35.5\\
STEm-Seg\cite{Athar_Mahadevan20ECCV}&ResNet-50&-&30.6&50.7&33.5&31.6&37.1\\
STEm-Seg\cite{Athar_Mahadevan20ECCV}&ResNet-101&2.1&34.6&55.8&37.9&34.4&41.6\\
MaskProp\cite{bertasius2020classifying}&ResNet-50&-&40.0&-&42.9&-&-\\
MaskProp\cite{bertasius2020classifying}&ResNet-101&-&42.5&-&45.6&-&-\\
\hline
\textbf{\Ours}&ResNet-50&30.0/69.9&36.2&59.8&36.9&37.2&42.4\\
\textbf{\Ours}&ResNet-101&27.7/57.7&40.1&64.0&45.0&38.3&44.9\\
\end{tabular}
\caption{\textbf{Video instance segmentation} AP (\%) on the YouTube-VIS \cite{vis2019} validation dataset. Note that,
for the first three methods, we have cited the results reported by the re-implementations in \cite{vis2019} for VIS. Other results are adopted from their original paper.
For the speed of \Ours we report the FPS results with and without the data loading process. 
Here we naively load the images serially, taking unnecessarily long time.
The data loading process can be much faster by parallelizing.
}
\label{tab:sota}
\vspace{-0.4cm}
\end{table*}

\begin{figure*}[h]
\centering
\includegraphics[width=\linewidth]{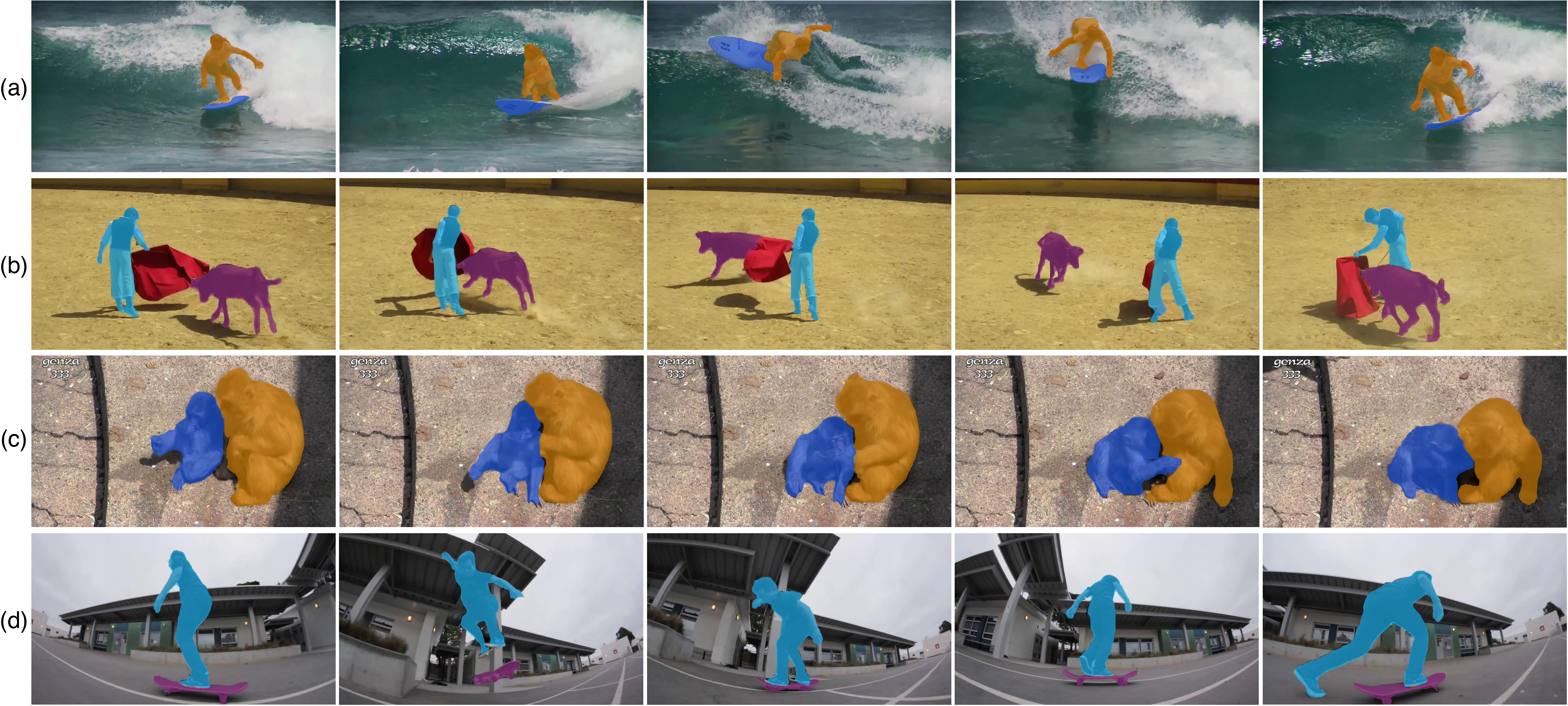}
\caption{\textbf{Visualization of \Ours} on the YouTube-VIS \cite{vis2019} validation dataset. Each row contains images from the same video. For each video, here the same colors 
depict the mask sequences of the same instances (Best viewed on screen).
}
\label{fig:3}
\vspace{-0.5cm}
\end{figure*}

We compare \Ours against some state-of-the-art methods in video instance segmentation in Table~\ref{tab:sota}. The comparison is performed in terms of both accuracy and speed. 
The methods in the first three rows are originally proposed for tracking or VOS.
We have cited the results reported by the re-implementations in 
\cite{vis2019} for VIS. Other methods including the MaskTrack RCNN, MaskProp \cite{bertasius2020classifying} and STEm-Seg \cite{Athar_Mahadevan20ECCV} are originally proposed for the VIS task in the temporal order. 

For the accuracy measured by AP, \textit{\Ours achieves the best result among methods using a single model without any bells and whistles}. Using the same backbone of ResNet-50 \cite{he2016deep}, \Ours achieves about 6 points higher %
in AP
than the MaskTrack R-CNN and the recently proposed STEm-Seg method. Besides, we argue the AP gap between \Ours and MaskProp mainly comes from its combination of multiple networks, \textit{i.e.}, Spatiotemporal Sampling Network \cite{bertasius2018object}, Feature Pyramid Network \cite{lin2017feature}, Hybrid Task Cascade Network \cite{chen2019hybrid} and the High-Resolution Mask Refinement post-processing. 
Since our aim is to design a conceptually simple and end-to-end framework, many improvements methods, such as complex video data augmentation and multi-stage mask refinement are beyond the scope of this work. For the speed measured by FPS (frames per second), \Ours shows a
significant 
advantage among all the reported results, achieving 27.7 FPS with the ResNet-101 backbone. If excluding the data loading process of multiple images, the speed 
can
achieve 57.7 FPS. Note that,
as we load the images in serial, the data loading process can be easily parallelized.
The fast speed of \Ours owes to its design of parallel decoding and no post-processing. 

The visualization of \Ours on the YouTube-VIS\cite{vis2019} validation dataset is shown in \figref{fig:3}, with each row containing images sampled from the same video. 
\Ours can track and segment instances well in challenging situations
such as: (a) instances overlapping, (b) changes of relative positions between instance, (c) confusion by the same category instances 
that are close together and (d) instances in %
various 
poses.

\section{Conclusion}
In this paper, we have proposed a new video instance segmentation framework built upon Transformers, which views the VIS task as a direct end-to-end parallel sequence decoding/prediction problem. In this way, instance tracking is achieved \textit{seamlessly and naturally} in the same framework of instance segmentation, which is significantly different
from and simpler than existing approaches, considerably simplifying the overall pipeline. 
Without bells and whistles, \Ours achieves 
the best result and the highest speed among methods using a single model on the YouTube-VIS dataset. 
To our knowledge, our work is the first one that applies the Transformer 
to video instance segmentation. We hope that similar approaches can 
be applied to many more video understanding tasks in the future.

\textbf{Acknowledgements
}
This work was in part supported by Beijing Science and Technology Project (No.\ Z1\-8\-1\-1\-0\-0\-0\-0\-8\-9\-1\-8\-0\-1\-8).
CS and his employer received no financial
support for the research, authorship, and/or publication of this article.

{\small
\bibliographystyle{ieee_fullname}
\bibliography{draft}
}

\end{document}